\documentclass[lettersize,journal]{IEEEtran}
\usepackage{amsmath,amsfonts}
\usepackage{algorithmic}
\usepackage{algorithm}
\usepackage{array}
\usepackage[caption=false,font=normalsize,labelfont=sf,textfont=sf]{subfig}
\usepackage{textcomp}
\usepackage{stfloats}
\usepackage{url}
\usepackage{verbatim}
\usepackage{graphicx}
\usepackage{cite}
\usepackage{multirow}
\usepackage{hhline}
\usepackage{booktabs}
\usepackage{makecell}
\usepackage{color}
\begin{document}

\title{A Training-Free Framework for Video License Plate Tracking and Recognition with Only One-Shot}

\author{
	Haoxuan~Ding,
    Qi~Wang,~\IEEEmembership{Senior~Member,~IEEE,}
	Junyu~Gao,~\IEEEmembership{Member,~IEEE,}
    and~Qiang~Li,~\IEEEmembership{Member,~IEEE,}

\thanks{H. Ding is with the Unmanned System Research Institute, and with the School of Artificial Intelligence, Optics and Electronics (iOPEN), Northwestern Polytechnical University, Xi’an 710072, P. R. China. (e-mail: haoxuan.ding@mail.nwpu.edu.cn)}%


\thanks{Q. Wang, J. Gao, and Q. Li are with the School of Artificial Intelligence, Optics and Electronics (iOPEN), Northwestern Polytechnical University, Xi'an 710072, P.R. China. (e-mail: crabwq@gmail.com, gjy3035@gmail.com, liqmges@gmail.com)}%

\thanks{Q. Wang is the corresponding author.}
}

\IEEEpubid{0000--0000/00\$00.00~\copyright~2021 IEEE}

\maketitle

\begin{abstract}
Traditional license plate detection and recognition models are often trained on closed datasets, limiting their ability to handle the diverse license plate formats across different regions. The emergence of large-scale pre-trained models has shown exceptional generalization capabilities, enabling few-shot and zero-shot learning. We propose OneShotLP, a training-free framework for video-based license plate detection and recognition, leveraging these advanced models. Starting with the license plate position in the first video frame, our method tracks this position across subsequent frames using a point tracking module, creating a trajectory of prompts. These prompts are input into a segmentation module that uses a promptable large segmentation model to generate local masks of the license plate regions. The segmented areas are then processed by multimodal large language models (MLLMs) for accurate license plate recognition. OneShotLP offers significant advantages, including the ability to function effectively without extensive training data and adaptability to various license plate styles. Experimental results on UFPR-ALPR and SSIG-SegPlate datasets demonstrate the superior accuracy of our approach compared to traditional methods. This highlights the potential of leveraging pre-trained models for diverse real-world applications in intelligent transportation systems. The code is available at \url{https://github.com/Dinghaoxuan/OneShotLP}.
\end{abstract}


\section{Introduction} \label{sec:intro}

\begin{figure}[h]
	\centering
	\subfloat{
		\label{fig0a}
		\includegraphics[width=0.49\linewidth]{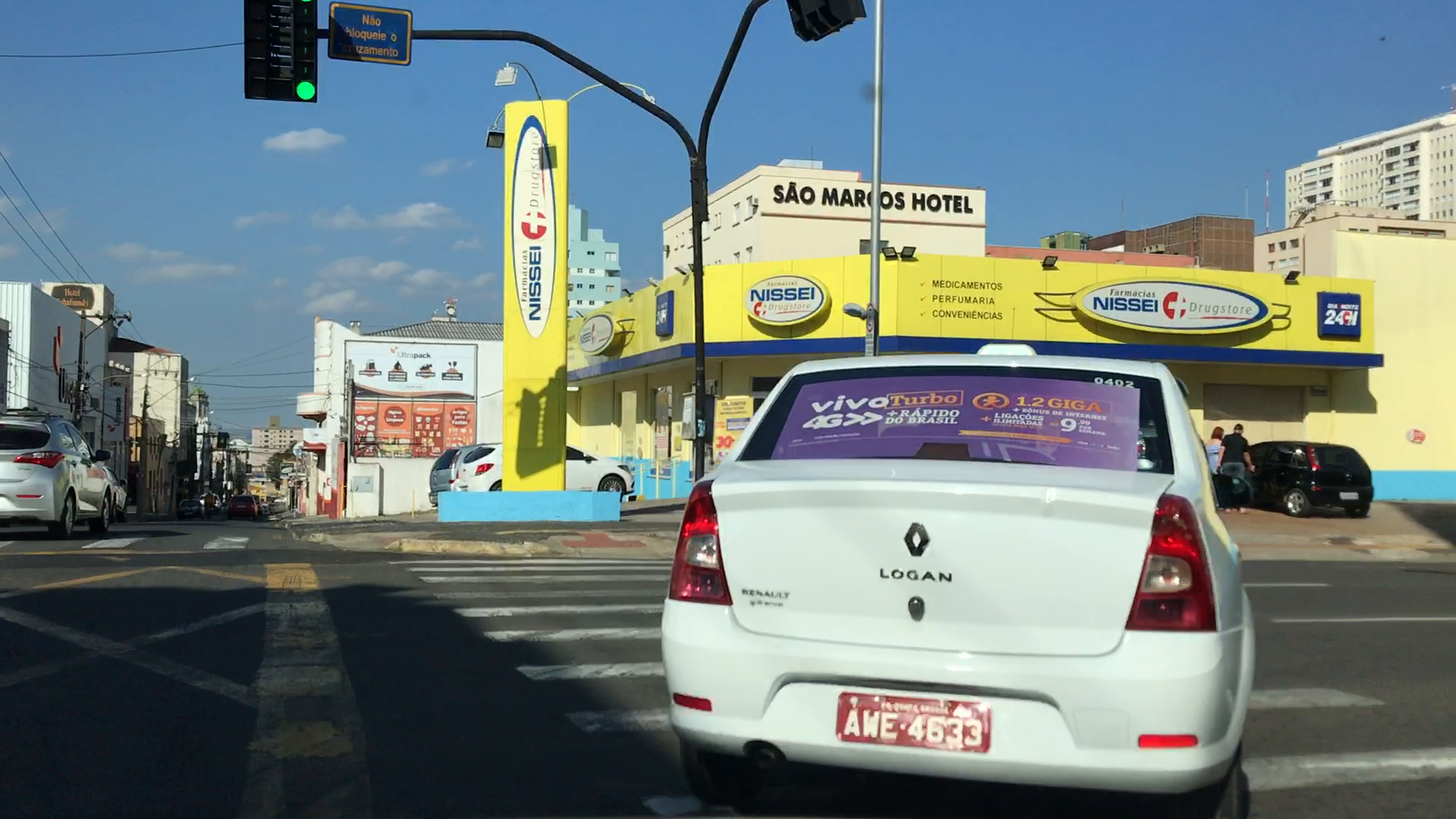}
		\includegraphics[width=0.49\linewidth]{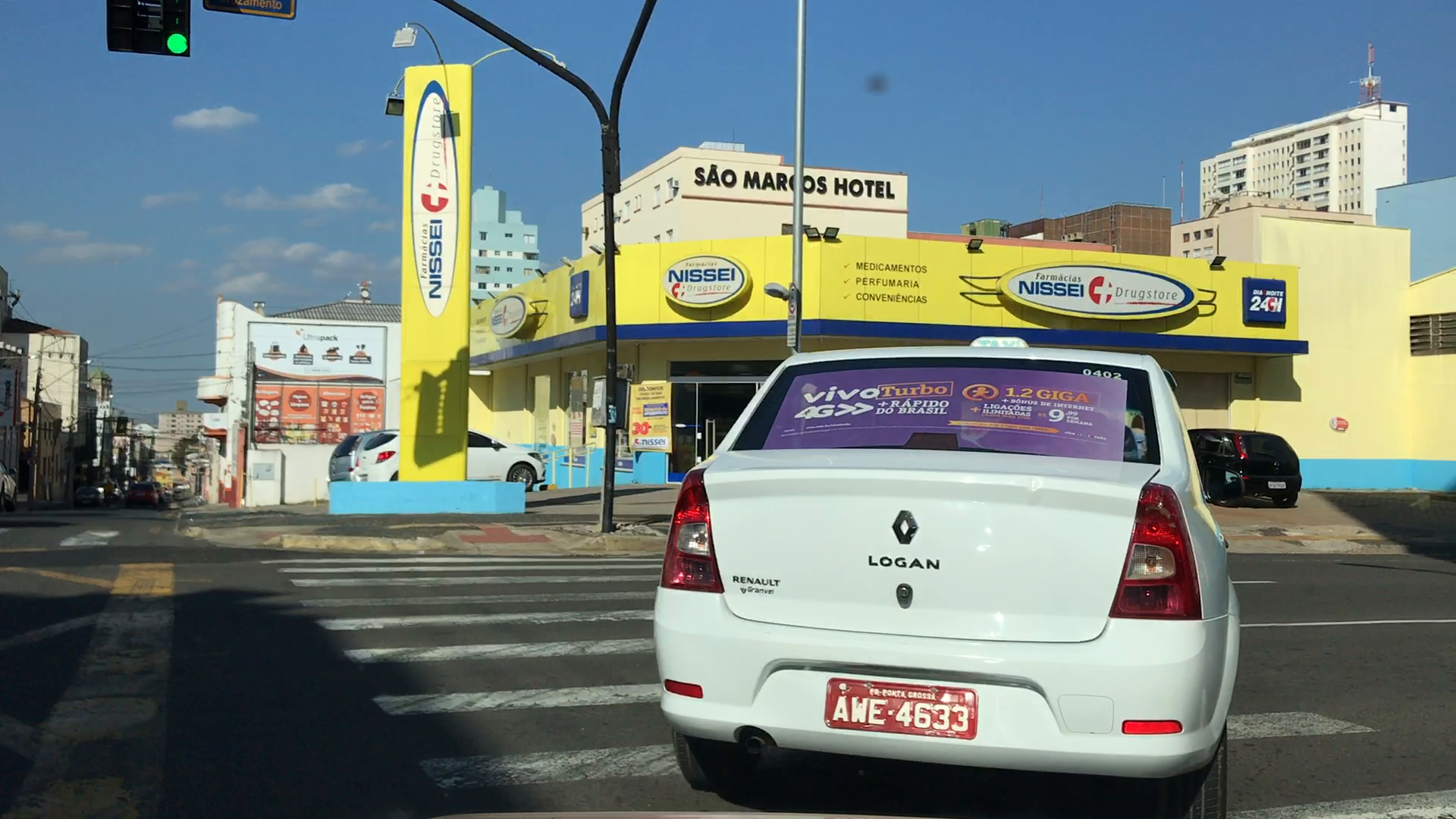}
	} \\
	\subfloat{
		\label{fig0b}
		\includegraphics[width=0.49\linewidth]{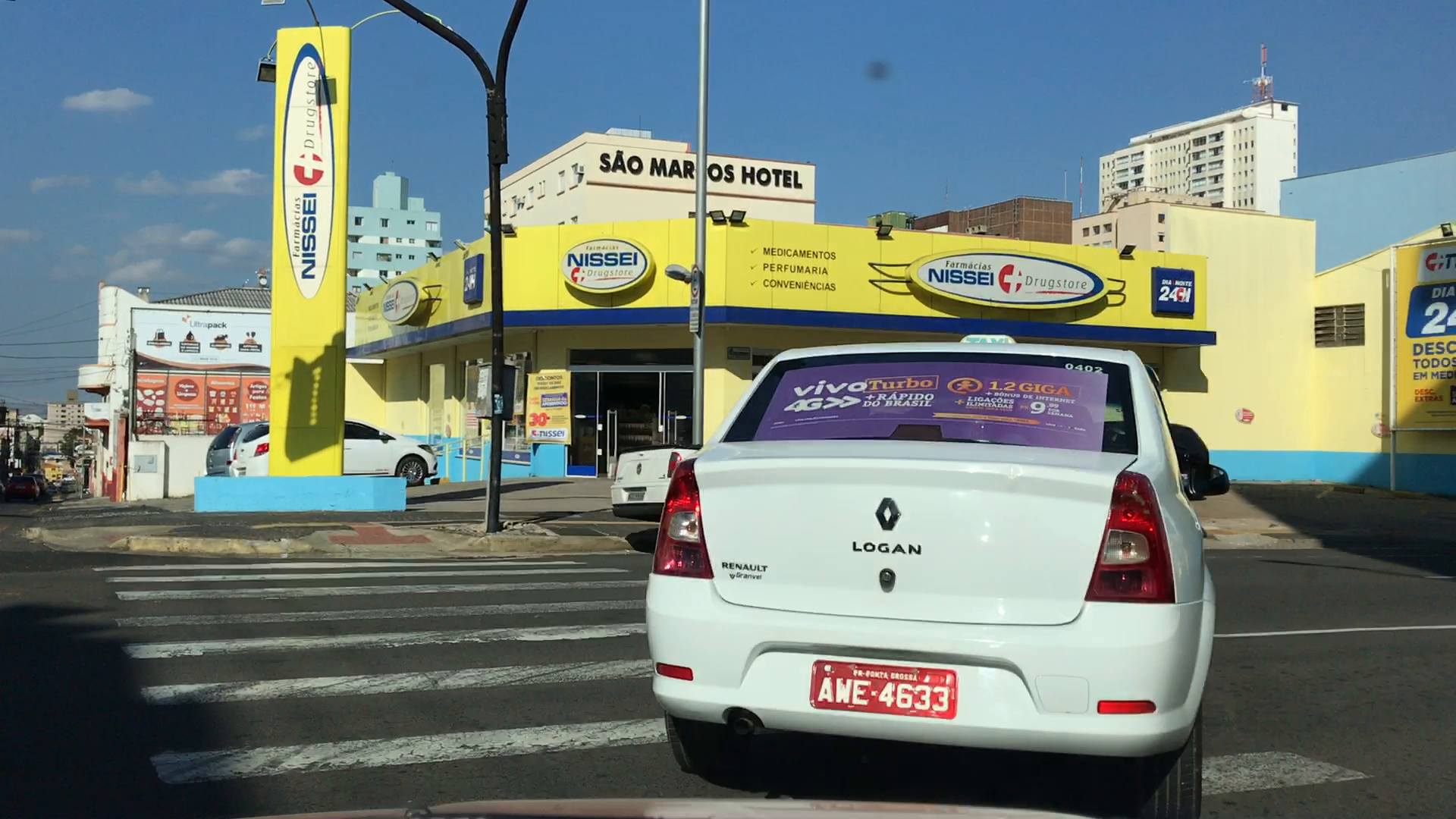}
		\includegraphics[width=0.49\linewidth]{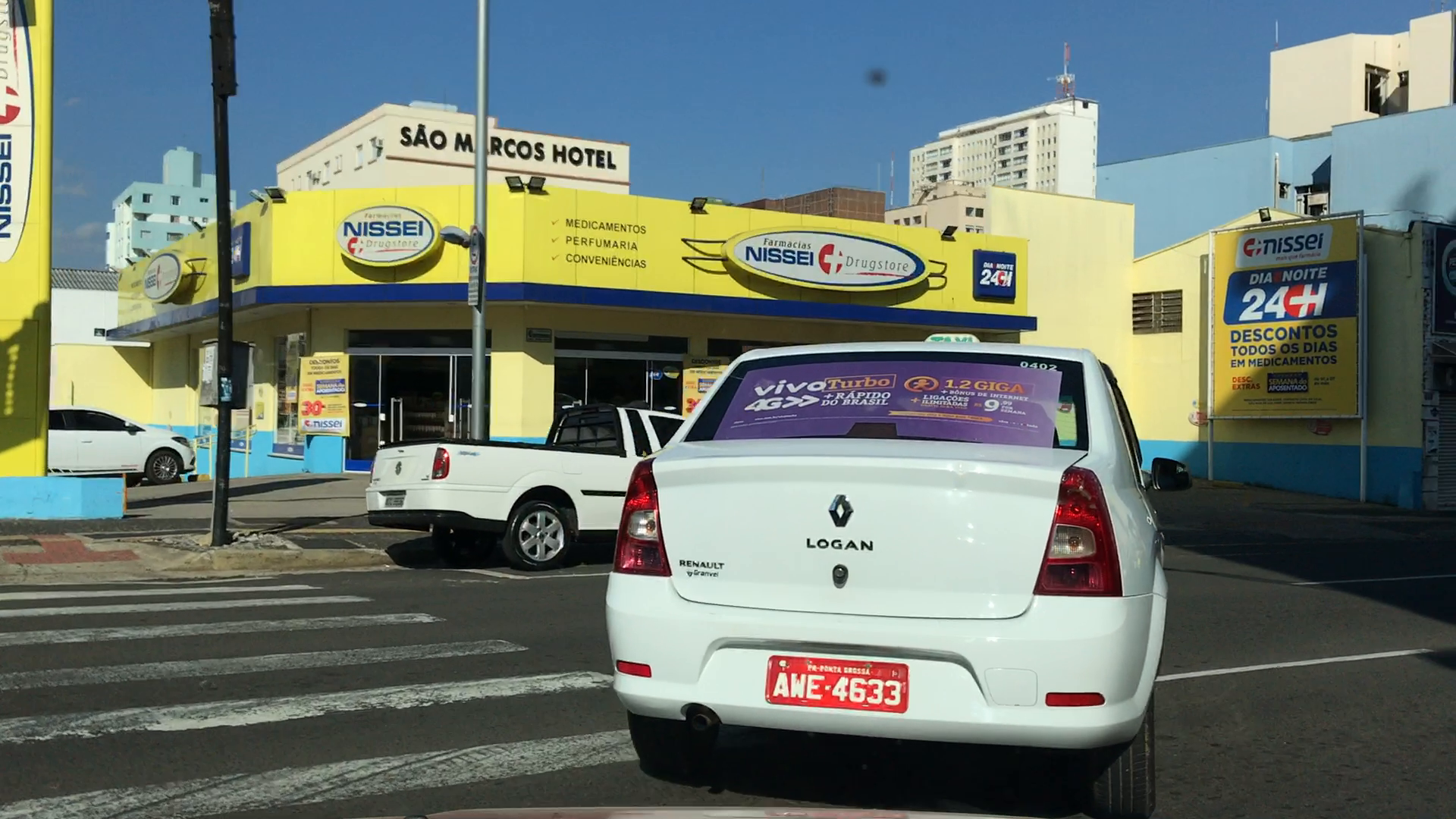}
	}
	\caption{The video clip with license plate information in transportation.}
	\label{fig0}
\end{figure}

\IEEEPARstart{L}{icense} plate (LP) serve as unique identifiers for vehicles, holding special significance within transportation. In dynamic traffic scenarios, automatic license plate tracking and recognition \cite{DBLP:journals/tits/LiuC19, DBLP:journals/tits/ZhangWLLSZ21, DBLP:journals/tits/Molina-MorenoGD19, DBLP:journals/tits/Al-ShemarryLA20,SCCA,CLPD} significantly facilitates the monitoring of vehicle driving behavior and supports the traffic management and safety of road systems. It is pivotal in the application of intelligent transportation systems (ITS). However, the diversity in license plates from different countries and regions poses a significant challenge for the deployment of automatic LP analysis system. The LP detection and recognition methods trained on a specific plate type struggle to adapt to the detection and recognition tasks for those plates with other types. Therefore, different countries and regions need to develop specialized license plate recognition systems to meet the requirements of traffic management, significantly increasing the costs associated with data collection, annotation, and model training. 

\IEEEpubidadjcol

Recently, Foundation models \cite{DBLP:journals/corr/abs-2108-07258} have emerged as a new paradigm in artificial intelligence, and they achieve promising performances in many tasks, including Natural Language Processing (NLP) \cite{GPT3, ChatGPT, GPT4}, Computer Vision (CV) \cite{SAM}. Notably, foundation models have demonstrated impressive performance in few-shot and zero-shot learning scenarios, where they achieve high accuracy with minimal or no task-specific data. This exceptional generalization capability of foundation models enable them to adapt to the detection and recognition among various LP types. The usage of foundation model make it possible to build a general LP analysis framework. And it significantly reduces the training burden for diverse license plate types, thereby lowering development costs and enhancing application efficiency. Thus, we propose \textbf{OneShotLP}, a training-free LP tracking and recognition framework based on foundation models for video LP analysis. The proposed OneShot framework operates without any need for training. By merely annotating the central position of the LP in the first frame of a video, the OneShotLP effectively execute continuous LP tracking and recognition across the entire video sequence. Benefiting from the zero-shot ability of introduced foundation models, the OneShotLP framework is capable of seamlessly adapting to the tracking and recognition of diverse LP types, eliminating the need for additional training or model modifications. The proposed framework makes it feasible to establish a universal license plate analysis system.

There are three core modules in OneShotLP, \emph{i.e.} tracking module, segmentation module, and recognition module. With the rapid development of point tracking \cite{om, DBLP:journals/corr/abs-2312-00786, CoTracker} in recent years, we introduce a point tracker \cite{CoTracker} into OneShotLP to track the annotated points across the whole video sequence. The point tracker provides an approximate location of LP instance annotated by point. This point provided by point tracker serves as a valuable cur for segmentation module. We select a promptable segmentation model, Segment Anything Model (SAM) \cite{SAM}, as segmentation module. To increase the analysis efficiency, we use the light-weight version of SAM \cite{SAM}, named EfficientSAM \cite{EfficientSAM} in proposed OneShotLP. The EfficientSAM segments potential targets within an image based on provided prompts (\emph{e.g.} point, box, and mask). Under the guidance from prompts, segmentation module precisely delineates and marks area of LP instance in complex visual environment, achieving robust LP detection and tracking. Meanwhile, to recognize license plate numbers from the detection result, we exploit a multimodal large language model (MLLM) \cite{LLaVA, QwenVL, Monkey, TinyLLaVA} as the recognition module in our framework. Leveraging the image comprehension and reasoning capabilities of multimodal large language model, the input image patch is analyzed autonomously by multimodal large language model to identify license plate numbers.

All these modules have excellent zero-shot learning performance, so that our proposed OneShotLP have the ability to directly adapt to the video transportation analysis of various types of LPs without the need for any training data. OneShotLP allows for immediate deployment across scenarios with different LP types, significantly reducing the resource expenditure typically required for training. In summary, the main contributions of this paper are:
\begin{enumerate}
  \item Propose a training-free LP tracking and recognition framework, OneShotLP, for video LP analysis. The proposed OneShotLP framework requires no training and only needs a single prompt point marking the approximate location of the LP in the first frame of video. Instructed by the prompt point, OneShotLP tracks the corresponding LP in the whole video sequence accurately and recognize the LP numbers by multimodal analysis. 
  \item Design a LP tracking and recognition pipeline for video LP analysis in OneShotLP. The pipeline consists three modules for LP location tracking, LP segmentation, and LP recognition, respectively. The tracking module track the prompt point from the first frame across the whole video sequence. And the potential segmentation mask at point is extracted by segmentation module. Finally, the segmented image patch is fed to recognition module to get the LP numbers via a multimodal model. 

  \item We combine foundation models in proposed OneShotLP framework for tracking, detection, and recognition. Their ability to generalize from limited inputs enables effective handling of complex and varied transportation scenarios without the need for extensive retraining. The experimental results show that our proposed OneShotLP has promising zero-shot learning performance, ensuring accurate and robust performance on video LP tracking and recognition.
\end{enumerate}

The rest if this paper is organized as follows. Section \ref{sec:related} provides a brief review of relevant issues in proposed OneShotLP framework. Section \ref{sec:method} elaborates in detail on our proposed OneShotLP framework. Section \ref{sec:exp} presents findings from several video license plate detection and recognition datasets, and analyzes factors influencing the process. Finally, Section \ref{sec:con} summarizes the research findings of this paper.

\section{Related Work}\label{sec:related}
In this section, We first provide an overview of emerging point tracking approaches in Section \ref{sec:related:point}. We then review researches on large visual models used for promptable segmentation and their applications in Section \ref{sec:related:sam}. And then, we introduce the relevant issues about multimodal large language models. Finally, we first summarize existing LP detection and recognition methods in Section \ref{sec:related:lp}.

\subsection{Point Tracking} \label{sec:related:point}
Our proposed OneShotLP first track the annotated point in the first frame to indicate the approximate postion of LP across the whole video sequence. The point tracking task is first defined in TAP-Vid benchmark \cite{TAP-Vid}. Harley \emph{et al.} \cite{PIPs} propose PIPs to track selected points in a fixed sliding window. Inspired by TAP-Vid \cite{TAP-Vid} and PIPs \cite{PIPs}, TAPIR \cite{TAPIR} introduces a matching stage and refine stage for feed-forward point tracking. Utilizing a simplified PIPs \cite{PIPs}, PointOdyssey \cite{PointOdyssey} proposes a benchmark for long-term point tracking. Meanwhile, CoTracker \cite{CoTracker} introduces a versatile and potent tracking algorithm that utilizes a transformer architecture to monitor points continuously across a video sequence. OmniMotion \cite{om} designs a quasi-3D canonical volume for video representing and considers point tracking as bijections in this volume space. These stable and robust point tracking methods provide useful information about motion to assist down-stream vision tasks, and we exploit the motion as cues in our proposed OneShotLP for LP location.

\subsection{Promptable Segmentation} \label{sec:related:sam}
The tracked point in video sequence serves as prompt for LP detection or segmentation. Kirillov \emph{et al.} \cite{SAM} propose the first vision foundation model for promptable segmentation, named Segment Anything Model. SAM \cite{SAM} predicts potential object masks in image accurately, guided by provided prompts, \emph{i.e.} points, boxes, or masks. SAM \cite{SAM} is applied among variety of downstream tasks. Zhang \emph{et al.} \cite{DBLP:journals/corr/abs-2304-13785} apply LoRA \cite{LoRA} to adapt SAM \cite{SAM} for segmenting medical images. Similarly, Wu \emph{et al.} integrate Adapter \cite{Adapter} to refine SAM \cite{SAM} specifically for medical imaging. In the realm of object tracking and detection, Tracking Anything \cite{DBLP:journals/corr/abs-2304-11968} combines SAM \cite{SAM} and XMem tracker \cite{DBLP:conf/eccv/ChengS22} for object tracking. Meanwhile, SAM-PT \cite{SAMPT} combine the point tracker with SAM \cite{SAM} to achieve accurate video object segmentation. And Chen \emph{et al.} \cite{DBLP:journals/corr/abs-2304-09148} transfer SAM \cite{SAM} into camouflage object detection. Ding \emph{et al.} \cite{SamLP} convert SAM \cite{SAM} into LP detection task via LoRA tuning \cite{LoRA} and show the promising performance of vision foundation model in detection task. SAM3D \cite{DBLP:journals/corr/abs-2306-02245} directly utilizes SAM \cite{SAM} for segmenting objects in bird’s eye view lidar images and pinpointing their locations in point clouds based on these segment results. Regarding image editing, Inpaint Anything \cite{DBLP:journals/corr/abs-2304-06790} uses SAM \cite{SAM} to delineate editing regions, which are then refined by an inpainting model. InpaintNeRF360 \cite{DBLP:journals/corr/abs-2305-15094} also employs SAM \cite{SAM} for segmenting regions in data prepared for NeRF training. Lastly, Zhang \emph{et al.} \cite{DBLP:journals/corr/abs-2305-03048} introduce a training-free one-shot method using SAM \cite{SAM} to locate specified objects in images and guide the generation of stable diffusion \cite{DBLP:conf/cvpr/RombachBLEO22}. In addition, several studies aim to enhance the segmentation efficiency of SAM \cite{SAM}. FastSAM \cite{DBLP:journals/corr/abs-2306-12156}, MobileSAM \cite{DBLP:journals/corr/abs-2306-14289}, and EfficientSAM \cite{EfficientSAM} concentrate on boosting inference speeds to facilitate the real-time application of SAM \cite{SAM}.

\subsection{Multimodal Large Language Models (MLLMs)} \label{sec:related:MLLM}
Leveraging the ability of large language models \cite{InstructGPT, ChatGPT, FLAN1, FLAN2, llama, llama2}, MLLMs aim to tackle tasks across vision and language. There are three types of methods to make large language models understand vision information, \emph{i.e.} Multimodal Instructing Tuning, Multimodal In-Context Learning, and Multimodal Chain-of-Though. Multimodal Instructing Tuning introduces instruction tuning into large language models. LLaVA \cite{LLaVA} first uses GPT-4 \cite{GPT4} to generate high quality instruction data including text description and visual question answers. And then, it converts visual tokens into visual-language tokens to ensure language model understand visual information. Qwen-VL \cite{QwenVL}, Otter \cite{otter}, MiniGPT-4 \cite{minigpt4}, and TinyLLaVA \cite{TinyLLaVA} also use this pipeline to fine-tune large language models. Multimodal In-Context Learning use prompt examples in context to guide the model learn novel tasks. VISPROG \cite{VISPROG} uses the in-context learning ability to invoke tools for visual reasoning. Chameleon \cite{Chameleon} also utilizes in-context examples to guide the leverage of various tools. Multimodal Chain-of-Though \cite{DBLP:journals/corr/abs-2304-07919, DBLP:conf/emnlp/HimakunthalaORH23, DBLP:journals/corr/abs-2305-02317, DBLP:journals/corr/abs-2302-00923, DBLP:conf/nips/ZhengYTZY23} bulids the chain-of-thought across different modalities. 

In summary, the visual understanding from MLLMs are more in line with the way humans perceive the real world, and MLLMs have excellent understanding and reasoning ability in diverse tasks. Thus, we decide to exploit this promising power from MLLMs in LP recogntion task, and we introduce a MLLM into our proposed OneShotLP.

\subsection{LP Detection and Recognition} \label{sec:related:lp}
Recent LP detection methods primarily evolve from object detectors. Initially, LP detection \cite{DBLP:journals/soco/RafiquePJ18, DBLP:conf/bmvc/DongHLLZ17, TE2E} employs two-stage detectors \cite{FasterR-CNN}, which have largely been supplanted by one-stage detectors \cite{DBLP:conf/avss/HsuACS17, UFPR-ALPR, DBLP:journals/corr/abs-1909-01754, DBLP:journals/tii/0001WHHSC021, DBLP:journals/ict-express/JamtshoRW21, shahidi2022deep} to enhance efficiency. Moreover, recent advancements aim to improve the accuracy and robustness of LP detectors. For instance, Silva and Jung \cite{DBLP:conf/eccv/SilvaJ18, DBLP:journals/tits/SilvaJ22} concentrate on LP detection in unconstrained scenarios. Similarly, Wang \emph{et al.} \cite{DBLP:journals/tits/WangBZC22} and Fan \emph{et al.} \cite{fan2022improving} focus on detecting the four corners of LPs to ensure precise detection. Chen \emph{et al.} \cite{DBLP:journals/ijon/ChenTMLYCY21} explore the correlation between vehicles and plates to augment detector robustness. Lee \emph{et al.} \cite{DBLP:journals/tits/LeeJKJP22} and Ding \emph{et al.} \cite{SCCA, CLPD} apply contrastive learning to accurately differentiate LPs from text in similar scenes. Furthermore, Chen and Wang \cite{DBLP:journals/tits/ChenW22} approach LP detection as a segmentation task. Ding \emph{et al.} \cite{SamLP} propose the first LP detector based on SAM \cite{SAM}, which also treats LP detection as segmentation task. In video LP detection, flow-guided methods \cite{Zhangcong, DBLP:conf/mir/LuYW21, LSVLP} remain prevalent though they suffer from low efficiency due to the intricate computations involved in optical flow.

As for LP recognition, there are two categories LP recognition methods in research, \emph{i.e.} single-type LP recognition and multi-type LP recognition. Single-type methods are limited to recognizing a specific type of LP and lack the flexibility to adapt to different plates. In contrast, multi-type methods are capable of identifying a diverse range of LP types. For single-type recognition methods, some works \cite{TE2E, DBLP:journals/ijon/Wang0QCD18, DBLP:journals/corr/abs-1806-10447} utilize convolutional neural networks (CNNs) for feature extraction and CTC-based decoder for alignment and loss calculation. Zhang \emph{et al.} \cite{DBLP:journals/tits/ZhangWLLSZ21} extract image features into a 1D visual sequence and apply a sequence attention module for LP character feature extraction. In addition, some methods \cite{DBLP:journals/corr/abs-1909-01754, DBLP:journals/jvcir/SilvaJ20} for multi-type LP recognition propose detecting or segmenting characters and then assembling the predictions into a text string. To avoid character segmentation and reduce labor-intensive for annotation, some attention mechanisms \cite{DBLP:conf/icdar/LiuCLYCY21, DBLP:journals/tits/XuZLLLS22} are designed to extract character features in 2D space. 

We design a robust tracking pipeline for stable LP tracking in complex environment in OneShotLP and a multi-modal large language model is introduced into OneShotLP to analyze the LP characters. The proposed method in this paper is suitable for both single-type and multi-type LP recognition without any training. 

\begin{figure*}[t]
	\centering
	\includegraphics[width=\linewidth]{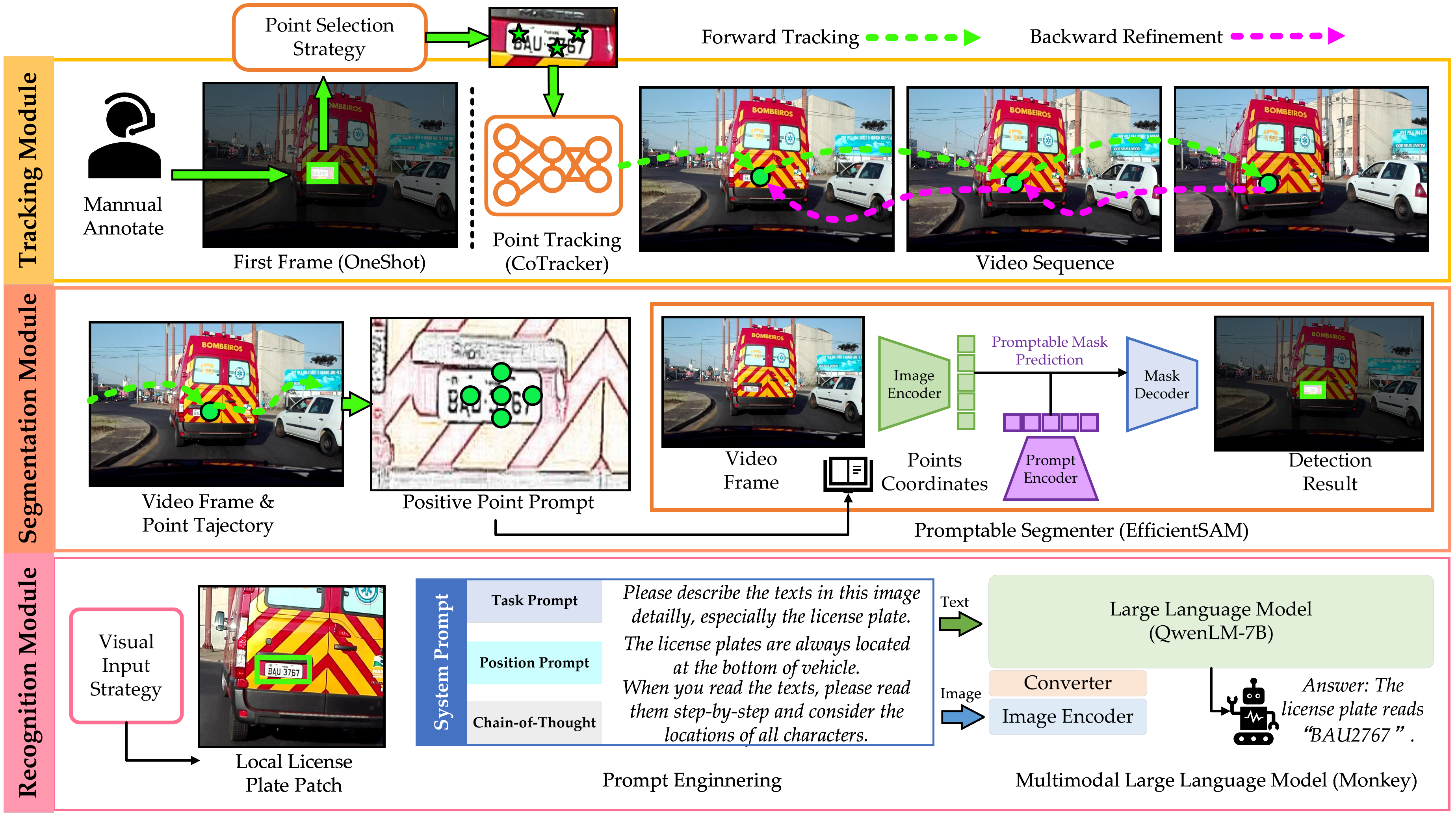}
	\caption{The pipeline of proposed OneShotLP.}
	\label{fig:pipe}
\end{figure*}

\section{Method} \label{sec:method}
In this section, we outline the specifics of our proposed OneShot framework. We begin with a brief overview of whole pipeline in Section \ref{sec:method:pipe}. And then, we detail the three core modules in OneShotLP. The tracking module employed in SamLP is described in Section \ref{sec:method:track}. In Section \ref{sec:method:sam}, we explain the promptable segmentation strategy for LP detection. In Section \ref{sec:method:mllm}, we describe the LP recognition module via multimodal large language model.

\subsection{Pipeline of OneShotLP}\label{sec:method:pipe}
As shown in Fig. \ref{fig:pipe}, the proposed OneShotLP is an automatic video LP analysis framework. It focuses on two main tasks, \emph{i.e.} LP detection and LP recognition. The tracking module and segmentation module are designed for LP detection task, and the recognition module concentrates on the LP recognition task. The inputs of OneShot is a video sequence $\mathcal{V}$ and annotated query points $q$ for LPs in the first frame of $\mathcal{V}$. The video sequence $\mathcal{V}$ and query points $q$ are first fed into tracking module. The tracking module first reinitializes the query points distribution in first frame via query point selection strategy. And then, tracking module predicts the trajectories of query points $q$ across whole frames in $\mathcal{V}$, denoted as $\mathcal{T}$. The query point trajectories $\mathcal{T}$ provide the prompt points $p$ for segmentation module. We use promptable segmentation model as segmentation module, and it outputs masks at the positions indicated by prompt points $p$. So far, the segmentation module predicts the masks for LPs in each video frame, achieving the LP detection task. After the detection of LPs, the local patches of LPs are input to recognition module, a multimodal large language model. The designed prompt question guides the recognition module to describe the content in image patches detailly. The characters on LP are obtained from those captions of LP patches, achieving the recognition of LPs.

\subsection{Tracking Module}\label{sec:method:track}


\textbf{Point Tracker:} We integrate the state-of-the-art CoTracker \cite{CoTracker} as the tracking module in OneShotLP framework. In traffic scenario, the dynamic environments impact the continuity of object visibility seriously. And CoTracker \cite{CoTracker} has proven highly effective, demonstrating substantial robustness in handling long-term tracking challenges such as object occlusion and the reappearance of objects. 

\textbf{Point Selection Strategy:} The process begins with the input of the video sequence $\mathcal{V}$ and the query points $q$. These points are strategically placed to identify the locations of LPs in the initial frame of $\mathcal{V}$. From this starting point, the tracking module is tasked with following these query points throughout the subsequent frames of the video. Thus, these query points directly influence the tracking accuracy. Some point selection strategies (Fig. \ref{fig:pointselection}) are attempted in tracking module to obtain tracking point $p$:

\begin{figure}[h]
	\centering
	\includegraphics[width=\linewidth]{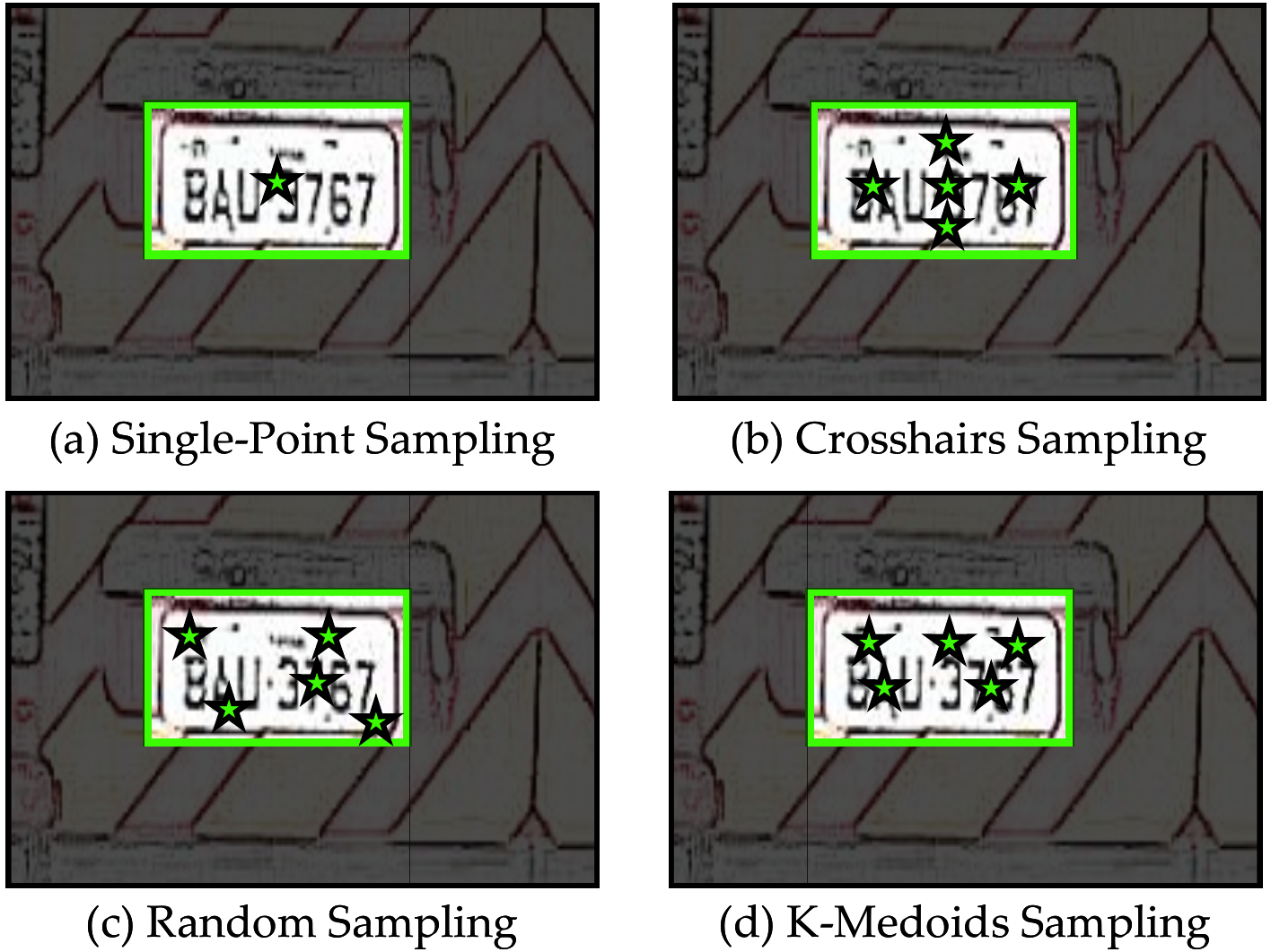}
	\caption{The point selection strategy to generate query points in tracking module.}
	\label{fig:pointselection}
\end{figure}

\begin{itemize}
    \item \textbf{Single-Point Sampling}: The annotated query point $q$ is treated as tracking point $p$ directly.
    \item \textbf{Crosshairs Sampling}: The offsets on four directions (\emph{i.e.} left, right, top, bottom) are added to query point $q$ to obtain five tracking points $p$, a single query point and four offset points.
    \item \textbf{Random Sampling}: The query point $q$ guides the segmentation model (explained in Section \ref{sec:method:sam}) to attain the LP mask in the first frame, and four tracking points $p$ are sampled from this LP mask randomly.  
    \item \textbf{K-Medoids Sampling}: Similarly to random sampling strategy, the LP mask is extracted under the prompt from query point $q$, and the tracking points $p$ are sampled from mask via K-Medoids clustering \cite{DBLP:journals/eswa/ParkJ09}. 
\end{itemize}

\textbf{Motion Trajectory:} Using CoTracker \cite{CoTracker}, we actively propagate these tracking points across all video frames to obtain motion trajectories $\mathcal{T}$. This propagation is not merely about tracking the points but also involves dynamically adjusting to changes in the environment that might obscure or temporarily remove the object from the frame. The module adeptly handles such scenarios, maintaining a coherent track throughout the sequence. As the points are tracked through the video, the tracking module generates detailed trajectories $\mathcal{T}$ that map the movement and changing positions of each point. These trajectories $\mathcal{T}$ provide instructions in the following promptalbe segmentation module for accurate LP localization.

\subsection{Segmentation Module}\label{sec:method:sam}

In the predicted trajectories $\mathcal{T}$ derived from our tracking module, the points in trajectories are crucial for promptable segmentation as they consistently indicate the location of the target object, \emph{i.e.} LPs, across the entire duration of the video sequence. To exploit these points for segmentation, we introduce a vision foundation model, Segment Anything Model (SAM) \cite{SAM, EfficientSAM}, as segmentation module in OneShotLP. SAM \cite{SAM} addresses a novel promptable segmentation task, where it operates by receiving a prompt (e.g., a point, box, or mask) and subsequently generating a segmentation mask at the location indicated by this prompt. This capability allows SAM \cite{SAM} to focus specifically on a user-defined area of an image, applying its segmentation algorithms precisely where needed. This method streamlines the process of segmenting objects within complex visual scenes, enhancing the accuracy of the segmentation task. Meanwhile, to balance the accuracy and efficiency, we use a lightweight version of SAM \cite{SAM}, \emph{i.e.} EfficientSAM \cite{EfficientSAM} as segmentation module in OneShotLP. 

EfficientSAM \cite{EfficientSAM} is an advanced iteration of the SAM \cite{SAM} that focuses on improving the computational efficiency and speed of the segmentation process without sacrificing accuracy. It is specifically designed to reduce the model's complexity and resource demands, enabling the effective segmentation on devices with limited computational power and in real-time applications. The core enhancement in EfficientSAM \cite{EfficientSAM} involves refining the feature extraction from tiny image encoder by knowledge distillation to ensure faster response times while maintaining high-quality segmentation results.

In each frame from $\mathcal{T}$, the image $\mathcal{I}$ is input to the image encoder of EfficientSAM \cite{EfficientSAM} to extract visual embeddings $\mathcal{E}_I$ of frame $\mathcal{I}$. Meanwhile, the visual embeddings $\mathcal{E}_I$ and the coordinates of prompt point $p$ (\emph{i.e.} the corresponding tracking point $p$ in trajectories $\mathcal{T}$) are fed to mask decoder of EfficientSAM \cite{EfficientSAM}. The mask decoder identifies potential high-response areas in $\mathcal{E}_I$ that correspond to the location of the prompt point $p$. It then decodes these areas to generate the object-level mask associated with that specific location. The object-level masks indicate the accurate locations of LPs in frame $\mathcal{I}$, achieving LP detection. 

Unlike traditional segment methods that typically require extensive training or fine-tuning on specific video segmentation datasets, the OneShotLP is a training-free framework and operates effectively in a zero-shot learning mode. This means that without any prior specific training on video data, our model can still perform exceptionally well in segmenting and tracking objects (\emph{i.e.} LPs). This is particularly advantageous in scenarios where the availability of annotated video segmentation data is limited or when quick deployment is necessary.

\subsection{Recognition Module}\label{sec:method:mllm}
To align the LP recognition approach more closely with human-like perception, we have incorporated a multimodal large language model into our OneShotLP framework. We replace the conventional LP recognition pipeline with a novel visual question answering approach. Leveraging the extensive pre-training and exceptional generalization capabilities of multimodal large language model, the recognition module in OneShotLP has the ability to adapt to a variety of LP styles without additional training, reaching higher stability and robustness in LP recognition and improving the effectiveness across diverse recognition scenarios. 

We selected Monkey-Chat \cite{Monkey} as our recognition module, which is a state-of-the-art MLLM on text visual question answering (VQA) task. Unlike other MLLMs, Monkey-Chat \cite{Monkey} has been specifically fine-tuned on diverse text datasets for recognition, VQA, and reasoning. Thus, Monkey-Chat \cite{Monkey} has developed exceptional text comprehension capabilities through fine-tuning which is suitable for LP number analysis. This is the reason why we choose this model (\emph{i.e.} Monkey-Chat \cite{Monkey}) as recognition module.

\begin{figure}[h]
	\centering
	\includegraphics[width=\linewidth]{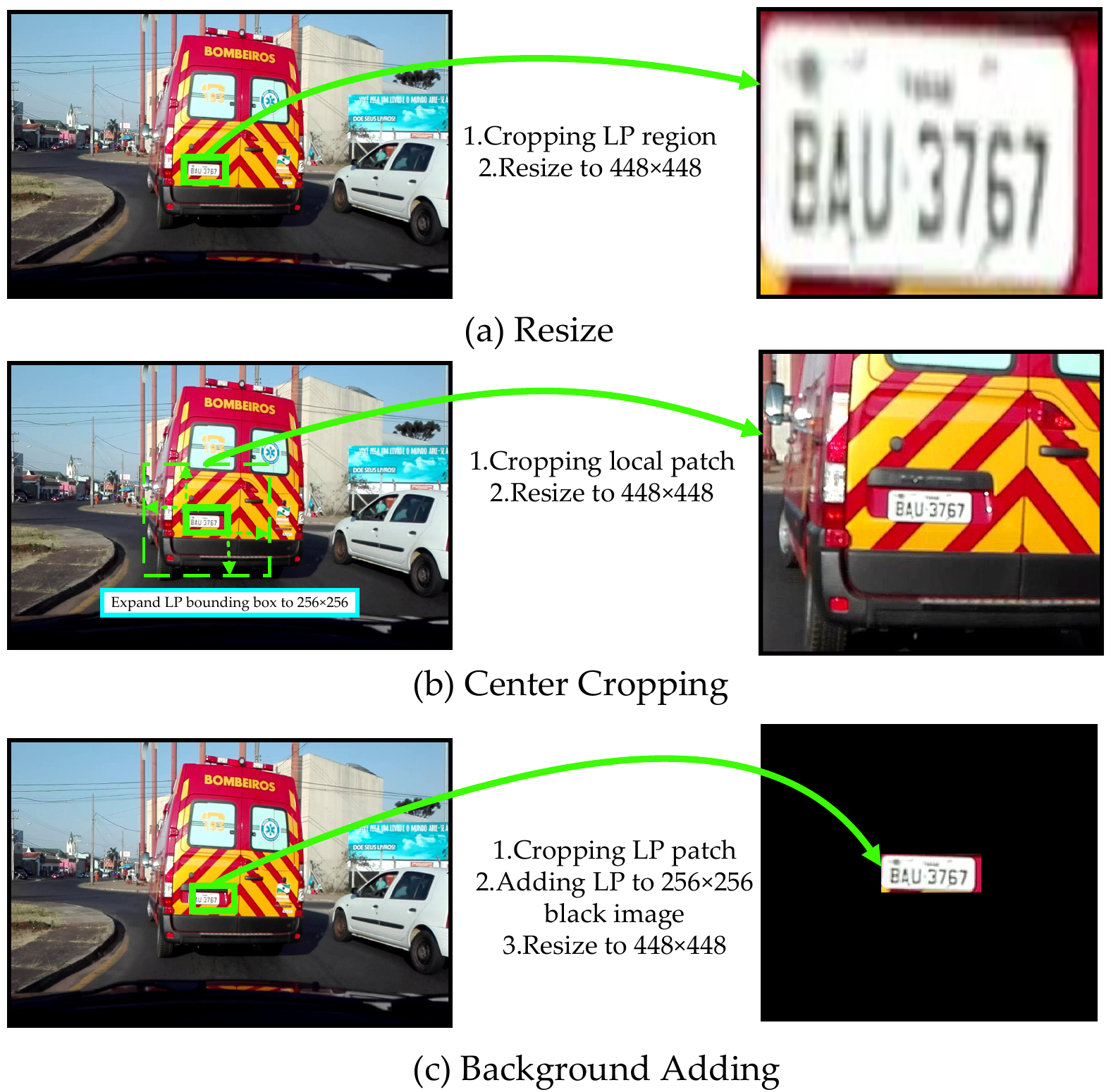}
	\caption{The input strategy for recognition module to achieve multimodal understanding and reasoning.}
	\label{fig:inputstrategy}
\end{figure}

However, the LP patch in street scene image is quite small and low-resolution, whereas the text images for MLLMs training are typically clear and high-resolution. Consequently, directly using MLLMs to analyze LP information from original street scene image frequently leads to incorrect responses and fails to produce accurate results. Thus, we exploit the predicted mask from segmentation module in recognition module to obtain local patch $\mathcal{I}_{LP}$ with LP. We attempt several visual input strategies (Fig. \ref{fig:inputstrategy}) to make $\mathcal{I}_{LP}$ match the input of MLLMs, obtaining better visual unstanding.

\begin{itemize}
    \item \textbf{Resize}: The local LP patch $\mathcal{I}_{LP}$ is directly resized to $448 \times 448$ and input to the image encoder.
    \item \textbf{Center Cropping}: The segmentation module provides the LP's location. Using the center point of LP mask as a coordinate, we perform a center cropping to extract a $256 \times 256$ local patch. This patch is then resized to $448 \times 448$ and used as input for the image encoder.
    \item \textbf{Background Adding}: The localized LP patch provided by the segmentation module is placed at the center of a $256 \times 256$ black background. Then, this combined image is resized to $448 \times 448$ as the input for the image encoder.
\end{itemize}

Meanwhile, we design a language prompt for Monkey-Chat \cite{Monkey} to instruct the model to answer the correct LP characters. Monkey-Chat \cite{Monkey} has not been trained on LP recognition datasets, so that directly asking the model \emph{“What is the license plate number?”} does not yield good results. Therefore, we convert the LP recognition task into a image captioning task, and the language prompts are discussed in Section \ref{sec:exp:dis_r}. 



\begin{table*}[h]
	\centering
	\caption{The comparison results of the proposed OneShotLP. The proposed OneShotLP achieves better performance compared to listed LP detection methods.}
	\begin{tabular}{m{0.18\linewidth} | c | m{0.14\linewidth}<{\centering} m{0.14\linewidth}<{\centering} m{0.14\linewidth}<{\centering} m{0.14\linewidth}<{\centering}}
		\toprule[1pt]
		Method  & Dataset  & P & R & F1-score & \multicolumn{1}{c}{AP} \\
		\midrule[1pt]
		OneShotLP (Ours) & UFPR-ALPR  & \textbf{98.3} & 98.7 & \textbf{98.5} & \textbf{97.5} \\
		\midrule[1pt]
		YOLOv2 \cite{YOLOv2} & \multirow{18}*{UFPR-ALPR}  &  92.7 & 94.7 & 93.7 & 87.8 \\
		YOLOv3 \cite{YOLOv3} &  & 94.9 & 97.4 & 96.1 & - \\
		YOLOv4 \cite{YOLOv4} &  & 93.1 & 92.6 & 92.8 & 88.5  \\
        YOLOv5 \cite{YOLOv5} & &- &- &- & 90.3  \\
		YOLOv7 \cite{YOLOv7} & &- &- &- & 90.9  \\
		YOLOv8 \cite{YOLOv8} & &- &- &- & 95.1  \\
		YOLOF \cite{YOLOF} & &- &- &- &  86.2  \\
		YOLOX \cite{YOLOX} & &- &- &- &  94.5 \\
		CentripetalNet \cite{CentripetalNet} & &- &- &- &	96.9 \\
		EAST \cite{EAST} &  & 92.8 & \textbf{99.9} & 96.2 & - \\
		R-FCN \cite{R-FCN} &  & 94.6 & 99.8 & 94.5 & - \\
		Laroca \emph{et al.} \cite{UFPR-ALPR} &  & - & 98.3 & - & - \\
		FGFA \cite{FGFA} &  & 97.2 & 98.3 & 97.7 & -  \\
        FCOS \cite{FCOS} \cite{FCOS} &   & 93.6 & 96.7 & 95.1 & 92.9 \\
        DFF \cite{DBLP:conf/cvpr/ZhuXDYW17} &  &- &- &- &  92.3 \\ 
		SELSA \cite{SELSA} &  &- &- &- & 93.2	\\ 
		Temporal RoI Align \cite{Temporal_ROI_Align} &  &- &- &- &  92.2 \\ 
		\midrule[1pt]
		\midrule[1pt]
        OneShotLP (Ours) & SSIG-SegPlate  & \textbf{96.9} & 99.5 & \textbf{98.2} & 97.6 \\
		\midrule[1pt]
		Silva and Jung \cite{DBLP:conf/sibgrapi/SilvaJ17} & \multirow{5}*{SSIG-SegPlate}   & 95.1 & 99.5 & 97.3 & - \\
		Laroca \emph{et al.} \cite{UFPR-ALPR} &  & - & \textbf{100.0} & - & - \\
		Silva and Jung \cite{DBLP:journals/jvcir/SilvaJ20} &  & 95.1 & 99.5 & 97.3 & - \\
		Laroca \emph{et al.} \cite{DBLP:journals/corr/abs-1909-01754} &  & 95.3 & 99.8 & 97.5 & - \\
        FCOS \cite{FCOS} &  & 96.4 & 99.9 & 98.1 & \textbf{97.7}  \\
		\bottomrule[1pt]
	\end{tabular}
	\label{tab:LPD}
\end{table*}

\section{Experiment} \label{sec:exp}

\subsection{Evaluation Metrics}\label{sec:exp:eval}

\textbf{LP Detection}: In object detection, three key metrics are used to evaluate performance: Precision (P), Recall (R), and F1-score. Precision measures the accuracy of the detected objects, calculated as the ratio of True Positive (TP) detections to the total number of positive detections, \emph{i.e.} True Positive (TP) and False Positive (FP). Recall assesses the model's ability to identify all relevant objects, defined as the ratio of True Positive (TP) detections to the total number of actual objects, \emph{i.e.} True Positive (TP) and False Negative (FN). F1-score is the harmonic mean of Precision and Recall, providing a balanced metric that considers both precision and recall.

\begin{equation}\label{eq:metric:det}
    \begin{split}
        &P = \frac{TP}{TP+FP}, \\
        &R = \frac{TP}{TP+FN}, \\
        &F1\mbox{-}score = \frac{2 \times R \times P}{P+R}.\\
    \end{split}
\end{equation}

\textbf{LP Recognition}: In LP recognition, Accuracy (Acc) is calculated as the ratio of correctly recognized LP instances to the total number of instances. This metric provides a straightforward measure of the model’s overall performance, indicating how often the model's predictions match the ground-truth.

\begin{equation}\label{eq:metric:reg}
        Acc = \frac{num_{TP}}{num_{GT}},
\end{equation}
where $num_{TP}$ is the number of correct predictions and $num_{GT}$ is the total number of ground-truths.

\subsection{Datasets}\label{sec:exp:data}
The proposed OneShotLP is evaluated on two video LP detection datasets, UFPR-ALPR \cite{UFPR-ALPR} and SSIG-SegPlate \cite{SSIG}.



\textbf{UFPR-ALPR} \cite{UFPR-ALPR}: This dataset is designed for video LP detection, specifically for Brazilian LPs. It includes $150$ video clips (totaling $4,500$ frames) captured from moving vehicles, with each clip consisting of $30$ frames. All frames have a consistent resolution of $1,920 \times 1,080$ pixels. The dataset is split into a training set and a testing set, each with $60$ videos, and a validation set with $30$ videos. Each video clip contains a single annotated license plate. The dataset features three types of license plates: gray LPs, red LPs, and motorcycle LPs.

\textbf{SSIG-SegPlate} \cite{SSIG}: This dataset focuses on Brazilian LP detection and comprises $2,000$ images taken with a fixed static camera. Every image has a resolution of $1,920 \times 1,080$ pixels and includes a single annotated LP. The dataset includes LPs from passenger vehicles, buses or trucks, and motorcycles. It is divided into three sets: $800$ images for training, $800$ images for testing, and $400$ images for validation.


\subsection{Implementation Details} \label{sec:exp:imp}
As a training-free framework, the proposed OneShotLP does not require complex additional configurations or hyperparameter adjustments, making it adaptable to various scenarios. This simplicity enhances its practicality and ease of use in diverse applications. The only requirement of OneShotLP is a single point annotation of the LP's center in the first frame of the video sequence. This annotation can be provided manually or generated by a LP detector. 

There are three main modules in OneShotLP. We select CoTracker \cite{CoTracker} as tracking module, EfficientSAM \cite{EfficientSAM} as segmentation module, and Monkey-Chat-7B \cite{Monkey} as recognition module. The video sequence is first input to CoTracker \cite{CoTracker} to capture the trajectory of annotated point. And then, EfficientSAM \cite{EfficientSAM} predicts the LP masks for each frame in video sequence guided by point trajectory. Finally, the local LP patch obtained by EfficientSAM \cite{EfficientSAM} and designed language prompt are fed to Monkey-Chat-7B \cite{Monkey} to get LP information. 

We only need to inference three modules in OneShotLP without any training, which dramatically reduce the consumption of GPU memory, making it possible to processing on a single NVIDIA RTX3090 GPU with 24GB memory. The framework is built by Hugging Face \cite{wolf-etal-2020-transformers} and Pytorch \cite{Pytorch}.

\subsection{Performance}\label{sec:exp:perf}

\textbf{LP Detection}: 
We evaluated the LP detection performance of our proposed OneShot method on the UFPR-ALPR \cite{UFPR-ALPR} and SSIG-SegPlate \cite{SSIG} datasets. The results demonstrate that our OneShot approach achieves better detection accuracy compared to mainstream LP detection methods on UFPR-ALPR \cite{UFPR-ALPR} (98.5 F1-score and 97.5 AP) and reach competitive accuracy on SSIG-SegPlate \cite{SSIG} (98.2 F1-score).

Notably, our proposed OneShot method does not require a large amount of training data; it can accomplish license plate detection across an entire video by providing the LP position in the first frame only. This approach leverages the generalization capability of pre-trained models, which endows it with strong few-shot learning abilities. By utilizing the provided one-shot data (\emph{i.e.}, the annotated first frame), our method effectively tracks the LP position throughout the video using point tracking techniques \cite{CoTracker}. Subsequently, the Segment Anything Model \cite{SAM, EfficientSAM} segments the LP region to obtain the bounding box.

\textbf{LP Recognition}: 

\begin{table}[h]
	\caption{The influence of system prompt for LP recognition.}
	\centering
	\begin{tabular}{l|c| m{0.18\linewidth}<{\centering} m{0.18\linewidth}<{\centering}}
		\toprule[1pt]
	  \multirow{2}*{Method} & \multirow{2}*{Dataset} & \multicolumn{2}{c}{Acc (\%)} \\
    \cline{3-4}
       & &correct $\geq$6 chars &correct$\geq$7 chars \\
        \midrule[1pt]
         OneShotLP (Ours) & UFPR-ALPR & \textbf{90.0} & \textbf{80.0} \\
		\midrule[1pt]
		Sighthound \cite{DBLP:journals/corr/MasoodSDO17}& \multirow{7}*{UFPR-ALPR} & 54.7 & 50.9 \\
        Sighthound \cite{DBLP:journals/corr/MasoodSDO17}$^{\star}$&  & 76.7 & 56.7 \\
        OpenALPR \cite{OpenALPR}&  & 54.7 & 50.9 \\
        OpenALPR \cite{OpenALPR}$^{\star}$&  & 73.3 & 70.0 \\
        Laroca \emph{et al.} \cite{UFPR-ALPR}&   & 87.3 & 64.9 \\
        Laroca \emph{et al.} \cite{UFPR-ALPR}$^{\star}$&  &  88.3 & 78.3 \\
        Gon\c{c}alves \emph{et al.} \cite{DBLP:conf/sibgrapi/GoncalvesDLMS18}& & - & 55.6 \\
        \midrule[1pt]
        \midrule[1pt]
        OneShotLP (Ours)& SSIG-SegPlate & 90.0 & 85.2  \\
        \midrule[1pt]
        Silva and Jung \cite{DBLP:journals/jvcir/SilvaJ20}& \multirow{4}*{SSIG-SegPlate} & 90.5 & 63.2 \\
        Laroca \emph{et al.} \cite{UFPR-ALPR}& & - & 85.4 \\
        Sighthound \cite{DBLP:journals/corr/MasoodSDO17}& & 89.0 & 73.1 \\
        OpenALPR \cite{OpenALPR}& & - & \textbf{87.4} \\
		\bottomrule[1pt]
	\end{tabular}
    \footnotesize{$^{\star}$ Redundancy method in \cite{UFPR-ALPR} is used in these methods.}
	\label{tab:LPR}
\end{table}

\begin{figure*}[h]
	\centering
	\includegraphics[width=\linewidth]{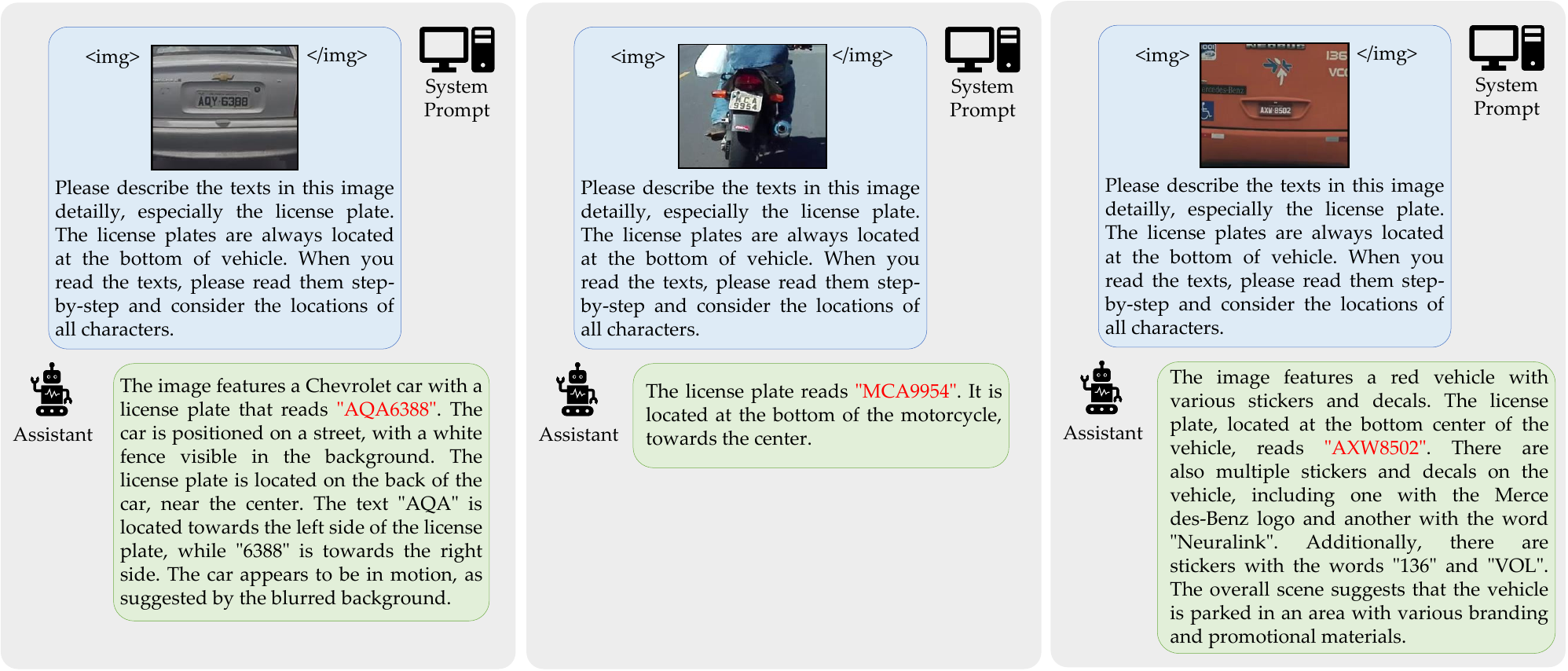}
	\caption{The visual question answering results for LP recognition.}
	\label{fig:vqa}
\end{figure*}

Experimental results demonstrate that leveraging a multimodal large language model (MLLM) for license plate recognition achieves competitive results (80.0 accuracy on UFPR-ALPR \cite{UFPR-ALPR}). However, there remains a performance gap compared to specialized LP recognition models. This disparity arises because current MLLMs have not been trained on LP recognition data and thus lack precise understanding of image details, leading to occasional hallucination effects. Despite these limitations, MLLMs exhibit a significant advantage in terms of generalization. They are capable of adapting to the recognition tasks of various LP styles, whereas specialized LP recognition models are typically trained on a specific style of license plate and often struggle to adapt to different styles. This flexibility makes MLLMs particularly valuable for applications requiring robust performance across diverse LP formats. Some answers from Monkey \cite{Monkey} for LP recognition are shown in Fig. \ref{fig:vqa}.

While MLLMs may currently fall short in accuracy compared to specialized models for LP recognition, their strong generalization capabilities present a promising avenue for further research and development. By addressing the issues related to image detail comprehension and reducing hallucination effects through fine-tuning of MLLMs, the performance of MLLMs can be significantly enhanced, potentially surpassing traditional OCR models in versatility and adaptability.

\subsection{Discussion About Detection}\label{sec:exp:dis_d}

\begin{table}[h]
	\caption{The influence of point selection strategies on UFPR-ALPR \cite{UFPR-ALPR}.}
	\centering
	\begin{tabular}{l|m{0.08\linewidth}<{\centering}|m{0.08\linewidth}<{\centering} m{0.08\linewidth}<{\centering} m{0.08\linewidth}<{\centering} m{0.08\linewidth}<{\centering}}
		\toprule[1pt]
	   Selection Strategy& \# of points & P & R & F1 & AP \\
		\midrule[1pt]
		Single-Point Sampling& 1 & 94.74 & 95.29 & 95.01 & 91.94 \\
        \midrule[1pt]
        \multirow{2}*{Crosshairs Sampling}& 5 & \textbf{98.28} & \textbf{98.68} & \textbf{98.48} & \textbf{97.53} \\
        & 11 & 39.42 & 42.70 &40.99 & 21.17 \\
        \midrule[1pt]
        \multirow{2}*{Random Sampling}& 5 & 95.30 & 95.52 & 95.41 & 92.52 \\
         & 11 & 95.08 & 95.46 & 95.27 & 92.33 \\
         \midrule[1pt]
        \multirow{2}*{K-Medoids Sampling}& 5 & 96.39 & 96.55 & 96.47 & 94.43 \\
        & 11 & 96.50 & 96.67 & 96.58 & 94.63 \\
		\bottomrule[1pt]
	\end{tabular}
	\label{tab:point}
\end{table}

\textbf{Point Selection}: As described in Section \ref{sec:method:track}, we attempt four query point selection strategies, \emph{i.e.} single-point sampling, crosshairs sampling, random sampling, and K-Medoids sampling, in tracking module. Table \ref{tab:point} presents the influence of different point selection strategies on various performance metrics. From this table, we find that only use a single point (\emph{i.e.} center point of LP) can not achieve robust tracking and LP detection. When we increase the number of query points in crosshairs sampling, random sampling, and K-Medoids sampling, the LP detection accuracy has significant improvement. The random sampling select points randomly, which may miss critical areas or introduce variability that affects consistency. K-Medoids sampling select points on LP area via clustering. The distribution of cluster center is sparse and dispersed, leading to misunderstanding of promptable segmentation. Therefore, crosshairs sampling provides aggregate distribution of tracking points, and these tracking points all focus on high-confidence LP instance, improving the stability of promptable segmentation in video sequence. 

\begin{table}[h]
	\caption{The influence of backward refinement on UFPR-ALPR \cite{UFPR-ALPR}.}
	\centering
	\begin{tabular}{l|m{0.08\linewidth}<{\centering} m{0.08\linewidth}<{\centering} m{0.08\linewidth}<{\centering} m{0.08\linewidth}<{\centering}}
		\toprule[1pt]
	  Propagation Method & P & R & F1 & AP \\
		\midrule[1pt]
		Only Forward & 98.17 & 98.68 & 98.42 & 97.47 \\
        Back Refinement & 98.28 & 98.68 & 98.48 & 97.53 \\
		\bottomrule[1pt]
	\end{tabular}
	\label{tab:back}
\end{table}

\textbf{Backward Refinement}: In tracking module, we introduce backward refinement to refine the trajectories. Specifically, we split point tracking into two loops. In the first loop, the query points $q$ from the first frame are tracked across entire video, and we obtain their locations in the last frame. In the second loop, we track these points reversely from the last frame to first frame to enhance the tracking stability and accuracy. In this case, we select crosshairs sampling method in OneShotLP according to the comparison above. The performance of backward refinement is shown in Table \ref{tab:back}. After introducing backward refinement, we observed an improvement in LP detection accuracy, indicating that backward refinement imposes robust consistency constraints on point tracking.

\subsection{Discussion About Recognition}\label{sec:exp:dis_r}

We designed three approaches to process the input license plate patches: Resize, Center Cropping, and Background Adding. The experimental results for these methods are presented in Table \ref{tab:mllminput}.

\begin{table}[h]
	\caption{The influence of input strategy for LP recognition on UFPR-ALPR \cite{UFPR-ALPR}.}
	\centering
	\begin{tabular}{l| m{0.25\linewidth}<{\centering} m{0.25\linewidth}<{\centering}}
		\toprule[1pt]
		\multirow{2}*{Input Strategy} & \multicolumn{2}{c}{Acc (\%)} \\
		\cline{2-3}
		&correct $\geq$6 chars&correct $\geq$7 chars \\
		\midrule[1pt]
		Resize & 50.0 & 33.3 \\
		Center Cropping & \textbf{90.0} & \textbf{80.0} \\
		Background Adding & 80.0 & 66.7 \\
		\bottomrule[1pt]
	\end{tabular}
	\label{tab:mllminput}
\end{table}

Directly resizing the LP patch to 448×448 pixels resulted in poor recognition performance. The low resolution of the LP patch, when scaled to the input size of the MLLM, caused the image to become blurry and indistinguishable, making it difficult for the MLLM to analyze the specific details of the license plate. In background adding strategy, the LP patch is placed at the center of a black background. The lack of contextual information in the surrounding environment hindered the MLLM's ability to relate the image content to relevant LP knowledge, resulting in lower recognition accuracy. The center cropping strategy preserves enough contextual information and maintains the recognizability of the LP after resizing. Consequently, center cropping achieved the best results among the three approaches.

In recognition module, we utilize a MLLM to understand and reason the LP numbers on LPs. A key factor influencing the performance of MLLMs is the formulation of prompts, which guide the model's output. Different prompts can lead to significantly varied results, highlighting the importance of prompt engineering in optimizing the system for specific tasks such as LP recognition. Prompt engineering involves systematically designing and refining the prompts provided to the MLLM to enhance its performance on targeted tasks. In our study, we explored various prompt formulations to identify the most effective system prompt for LP recognition. By analyzing the model's responses to different prompts, we aimed to determine how specific prompt characteristics influence the model's ability to accurately and consistently recognize LP numbers. The influences of system prompt for LP recognition are shown in Table \ref{tab:prompt}

\begin{table}[h]
	\caption{The influence of system prompt for LP recognition on UFPR-ALPR \cite{UFPR-ALPR}.}
	\centering
	\begin{tabular}{l| m{0.08\linewidth}<{\centering} m{0.08\linewidth}<{\centering}}
		\toprule[1pt]
	  \multirow{2}*{System Prompt} & \multicolumn{2}{c}{Acc (\%)} \\
    \cline{2-3}
        &\makecell{correct \\ $\geq$6 \\chars}&\makecell{correct \\ $\geq$7\\ chars} \\
        \midrule[1pt]
        \textit{What is the license plate number?} & 78.3 & 71.7 \\
		\midrule[1pt]
		\textit{What is the text on this licesne plate?} & 75.0 & 68.3 \\
        \midrule[1pt]
        \makecell[l]{\textit{Please describe the texts in this image } \\ \textit{step-by-step, especially the license plate.}} & 83.3 & 76.7 \\
        \midrule[1pt]
        \makecell[l]{\textit{The license plates are always located at the } \\ \textit{bottom of vehicle. Please describe the texts in this} \\ \textit{image step-by-step, especially the license plate.}}& 81.7 & 76.7 \\
        \midrule[1pt]
        \makecell[l]{\textit{Please describe the texts in this image detailly,} \\ \textit{especially the license plate. When you read the texts,} \\ \textit{please read them step-by-step and consider the} \\ \textit{locations of all characters.}}& 86.7 & 76.7 \\
        \midrule[1pt]
        \makecell[l]{\textit{Please describe the texts in this image detailly,} \\ \textit{ especially the license plate. The license plates} \\ \textit{are always located at the bottom of vehicle. When} \\ \textit{ you read the texts, please read them step-by-step} \\ \textit{ and consider the locations of all characters.}}& 90.0 & 80.0 \\
		\bottomrule[1pt]
	\end{tabular}
	\label{tab:prompt}
\end{table}

\textbf{Task Prompt}: We first directly input the task to MLLM with the prompt (\emph{i.e.}, \textit{What is the license plate number?} and \textit{What is the text on this licesne plate?}). We observed that the performance of Monkey \cite{Monkey} on LP recognition tasks was suboptimal when directly applied without specific instruction tuning for this task. The  Monkey \cite{Monkey}, although highly capable in various OCR tasks, has not been explicitly trained or fine-tuned on datasets specifically tailored for LP recognition. This lack of instruction tuning means that the model is not adequately prepared to handle the unique challenges and nuances associated with identifying and interpreting LP text (only reaching 71.7\% and 68.3 accuracy when corrected chars $\geq$ 7). Therefore, it is imperative to leverage prompt engineering to guide the MLLM in adapting to the novel task of license plate recognition. By crafting specific and well-structured prompts, we can effectively direct the MLLM’s attention towards the unique characteristics and requirements of LP recognition.

In designing the prompts for MLLM to enhance intelligent LP recognition, we considered various critical factors to optimize performance. Given that most MLLMs are primarily trained on captioning tasks, we adjusted our approach by not directly querying the LP numbers. Instead, we instructed the MLLM to provide a detailed description of the image's content, with particular emphasis on the LP information (\emph{i.e.}, \textit{Please describe the texts in this image step-by-step, especially the license plate.}). This method leverages the model's inherent strengths developed through extensive training on image caption datasets, and converts LP recognition task to image caption task, thereby enhancing its adaptability to the task of LP recognition. By framing the prompt to request a comprehensive description, the model is encouraged to engage more deeply with the visual details, ensuring that the LP features are thoroughly analyzed and articulated. This strategy capitalizes on the model’s pre-existing capabilities, fostering a more effective alignment with the new task requirements without necessitating extensive retraining. Experimental results demonstrate that this approach significantly improves the accuracy of LP recognition (achieving 76.7\% accuracy on 7 chars and 83.3\% on 6 chars).

\textbf{Chain of Thought Prompt}: To fully leverage the MLLM's reasoning capabilities for extracting license plate information, we introduced a chain-of-thought approach. This method involves guiding the MLLM to think step-by-step through prompts, ultimately arriving at the correct result (\emph{i.e.}, \textit{When you read the texts, please read them step-by-step and consider the locations of all characters.}). By structuring the prompts in a manner that encourages sequential reasoning, the model can better analyze and interpret the license plate data. Our experiments demonstrated that this chain-of-thought prompting significantly improved the accuracy of recognizing LPs, particularly for those with more than six correct characters (from 83.3\% accuracy to 86.7\% accuracy). This improvement underscores the effectiveness of breaking down the task into smaller, manageable steps, allowing the model to systematically process the information and reduce errors.


\textbf{Position Prompt}: To further enhance the accuracy of license plate recognition, it is crucial to ensure that the MLLM focuses on the information in the license plate region of the image. To address this issue and direct the model's attention to the LP region, we introduced position prompts (\emph{i.e.}, \textit{The license plates are always located at the bottom of the vehicles.}). These prompts convey common-sense information about the typical locations of license plates within images, guiding the model to extract relevant information from the corresponding areas. Experimental results demonstrate that incorporating position prompts leads to a significant improvement in recognition accuracy (reaching 80.0\% accuracy on 7 chars and 86.7\% on 6 chars). This approach ensures that the model more effectively attends to the critical regions of the image, thereby enhancing its ability to accurately identify LP numbers.


\section{Conclusion} \label{sec:con}
This paper introduced an innovative framework for the detection and recognition of license plates in video sequences, leveraging the strengths of multimodal large language models. By initiating with the position of the license plate in the first frame and employing a point tracking module, we efficiently tracked the license plate throughout the video. The generated position prompts were effectively used by a segmentation module to extract the license plate region, which was subsequently recognized by MLLMs.

The experimental results on UFPR-ALPR \cite{UFPR-ALPR} and SSIG-SegPlate \cite{SSIG} datasets highlight the efficacy of our OneShot method, demonstrating superior detection accuracy compared to traditional methods. Notably, our approach requires minimal training data, relying on the robust generalization capabilities of pre-trained models. This enables our framework to adapt to various license plate styles, addressing a significant limitation of conventional OCR models which often struggle with different license plate formats.

Future work will focus on further enhancing the accuracy and robustness of the model, particularly in challenging environments with varying lighting and occlusion conditions. Additionally, we aim to explore the integration of this framework into real-time systems for intelligent transportation applications, providing a scalable and efficient solution for license plate recognition.

\bibliography{ref}
\bibliographystyle{IEEEtran}

\end{document}